\documentclass[10pt,twocolumn]{article}

\usepackage[margin=1in]{geometry}
\usepackage{times}
\usepackage{authblk}
\usepackage{graphicx}
\usepackage{amsmath,amssymb,amsfonts}
\usepackage{booktabs}
\usepackage{multirow}
\usepackage{url}
\usepackage[hidelinks]{hyperref}
\usepackage{caption}
\usepackage{subcaption}
\usepackage{microtype}
\graphicspath{{figs/}{figures/}{pictures/}{images/}{./}}

\title{\bfseries Creating User-steerable Projections with Interactive Semantic Mapping}

\author[1]{A. A. Oliveira}
\author[1]{M. Espadoto}
\author[1]{R. Hirata Jr.}
\author[1]{R. M. Cesar Jr.}
\author[2]{A. C. Telea}

\affil[1]{Institute of Mathematics and Statistics, University of S\~ao Paulo \\\texttt{\{arturao, mespadot, hirata, cesar\}@ime.usp.br}}
\affil[2]{Department of Information and Computing Sciences, Utrecht University \\\texttt{a.c.telea@uu.nl}}

\date{}  

\begin{document}
	
	\maketitle

	\begin{abstract}
		Dimensionality reduction (DR) techniques map high-dimensional data into lower-dimensional spaces. Yet, current DR techniques are not designed to explore semantic structure that is not directly available in the form of variables or class labels. We introduce a novel user-guided projection framework for image and text data that enables customizable, interpretable, data visualizations via zero-shot classification with Multimodal Large Language Models (MLLMs). We enable users to steer projections dynamically via natural-language guiding prompts, to specify high-level semantic relationships of interest to the users which are not explicitly present in the data dimensions. We evaluate our method across several datasets and show that it not only enhances cluster separation, but also transforms DR into an interactive, user-driven process. Our approach bridges the gap between fully automated DR techniques and human-centered data exploration, offering a flexible and adaptive way to tailor projections to specific analytical needs.
	\end{abstract}
	
	
\begin{figure*}[!t]
	\centering
	\includegraphics[width=0.9\linewidth]{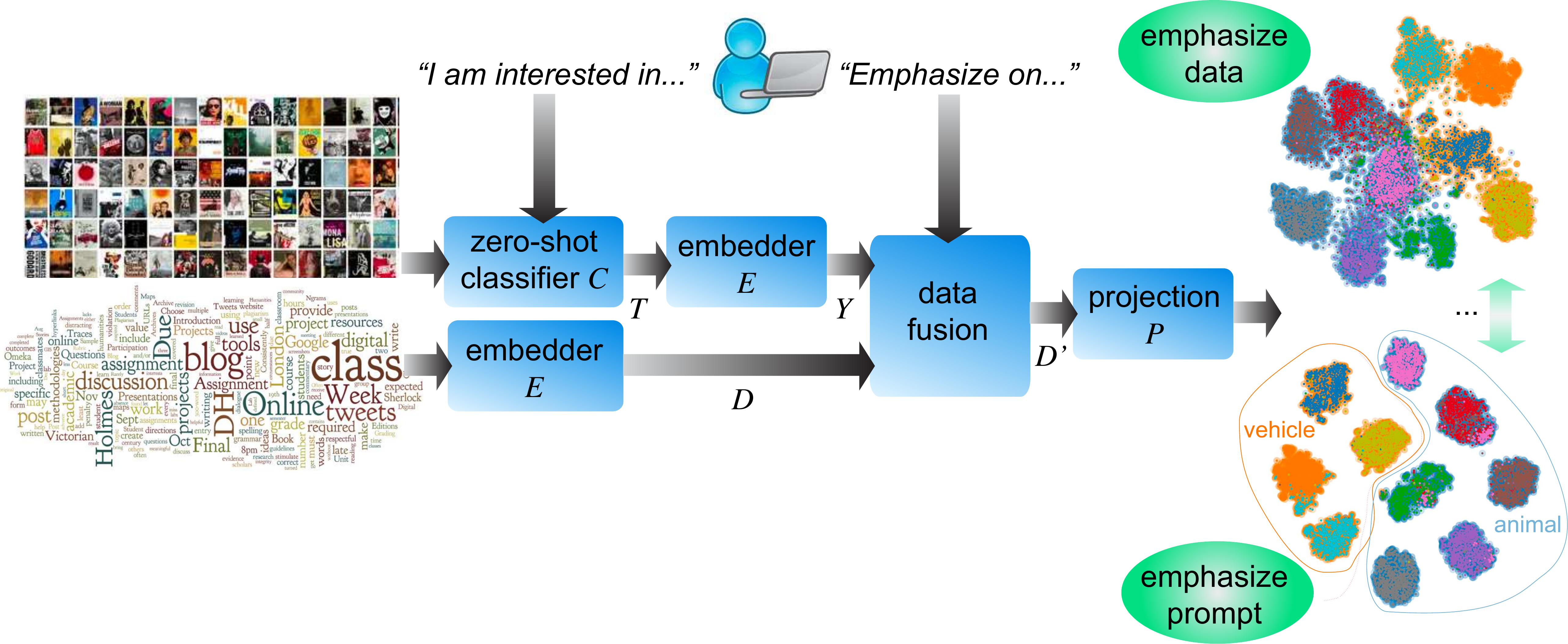}
	\caption{Given an image or text dataset, we create user-steerable projections by the following pipeline. We embed the input using the CLIP embedder $E$. In parallel, we analyze the input based on user-supplied prompts using the zero-shot Qwen classifier $C$. Its textual outputs $T$ are mapped to the same space as the input by the embedder $E$. Next, we fuse the embedded classifier results $Y$ and input embedding $D$ based on the user-desired emphasis. The result $D'$ is then projected using standard dimensionality reduction.}
	\label{fig:teaser}
\end{figure*}

	\section{Introduction}
	\label{sec:intro}
	Dimensionality reduction (DR) techniques, also known as multidimensional projections, are powerful tools for the visual exploration of high-dimensional datasets. Briefly put, given a dataset $D$, a projection $P(D)$ is a (typically two-dimensional) scatterplot which maps every data sample in $D$ to a scatterplot point so that the structure of $D$ is encoded into visual structures in $P(D)$. Many projection techniques exist which aim to preserve different data structure aspects such as global inter-point distances, point neighborhoods, or clusters of similar samples, to mention just a few such aspects of interest\,\cite{nonato18,espadoto19}. 

A particular use-case for projections involves the exploration of image and text datasets. In contrast to so-called tabular data, which have relatively few dimensions that explicitly capture orthogonal aspects of the studied problem, image and text data are usually described by a large number of dimensions (hundreds or more). These dimensions -- such as the colors of an image's pixels or latent features extracted from the data by Machine Learning (ML) techniques -- do not have a specific significance when taken separately and/or are hard to be directly interpreted by humans. Projecting such very high-dimensional datasets to yield visual structures which are useful for data analysis is a challenging problem, particularly when the data is very sparse\,\cite{bunte11,sorzano14_survey,espadoto19}. 

At an even higher level, projections typically create a \emph{single view} of what is, essentially, complex and high-dimensional data. It is hard to expect that all aspects of interest present in such data will be captured equally well in such a single view. So, how can users control what a projection captures? Typical mechanisms for this include preprocessing the data dimensions (\emph{e.g.} by fine-tuning their weights) or varying the projection's hyperparameters. However, both techniques are hard to use in practice -- users typically have high-level concerns or questions about their data that they would like a projection to emphasize; weights and hyperparameters typically control low-level aspects of a projection.

In this paper, we propose a method to steer the grouping of data points in the projected space. Our goal is to enable users to generate DR scatterplots that are easier to analyze to detect structures of interest and, in the same time, express what these structures of interest are in simple language, without the need to tune complex hyperparameters. We do this by leveraging the capabilities of ML models such as Qwen\,\cite{bai2025qwen25vltechnicalreport} and CLIP\,\cite{radford2021learning}, which can label data points using zero-shot classification and, more importantly, can embed semantically-related text and image data into the same latent space.

We show that projecting data embeddings fused with the embeddings of semantically meaningful text labels generates scatterplots that have better cluster separation than when using the data only and, also, that data gets grouped according to the provided label. Several types of labels can be used in this approach, from those obtained by using a zero-shot classifier and a set of possible classes to more sophisticated ones based on semantic concepts that exist, but are not explicitly encoded, in the data itself. For example, when using a digit dataset like MNIST\,\cite{lecunmnist2010}, instead of labeling the data according to the digits themselves, users can ask a model to label digits into well- or poorly-scribbled ones, by using the ability of MLLM's to understand these concepts. The same idea can be used to incorporate an extra layer of user-provided labels to group existing ones, thereby creating visualizations showing hierarchically-grouped data, such as digits grouped by similarity and then by parity (odd \emph{vs} even digits), as shown in Figure~\ref{fig:exp2}.

We structure this paper as follows: Section~\ref{sec:background} discusses related work on dimensionality reduction. Section~\ref{sec:method} details our method. Section~\ref{sec:experiments} presents our experimental setup alongside the results that support our contributions outlined above. Section~\ref{sec:discussion} discusses our proposal. Finally, Section~\ref{sec:conclusion} concludes the paper.

	\section{Background}
	\label{sec:background}
	Let $O=\{\mathbf{o}_i\}, 1 \leq i \leq N$ be a dataset containing $N$ observations of image or text data. An embedder model is a function $E(\mathbf{o}_i) \subset  \mathbb{R}^n$ which takes images or texts $\mathbf{o}_i$ as input and outputs $n$-dimensional embeddings $\mathbf{x} = (x^1,\ldots,x^n)$, $x^i \in \mathbb{R}, 1 \leq i \leq n$. Let $D=\{\mathbf{x}_i = E(\mathbf{o}_i) | \mathbf{o}_i \in O \}$ be a dataset of all embeddings of elements in $O$. $D$ can be seen as a table with $N$ rows (observations) and $n$ columns (dimensions). A dimensionality reduction or projection technique is a function
\begin{equation}
P : \mathbb{R}^n \rightarrow \mathbb{R}^q
\label{eqn:projdef}
\end{equation}
where $q \ll n$. We next focus on the typical case $q=2$. Projecting a dataset $D$ yields thus a 2D scatterplot which we denote next as $P(D)$.

A zero-shot classifier is a function $C(\mathbf{o}_i, p) = \mathbf{t}_i$ which takes a sample $\mathbf{o}_i \in O$ as input, and a prompt (or set of parameters) $p$, and outputs textual labels $\mathbf{t}_i$. Finally, yet $Y=\{\mathbf{y}_i \}$ be the dataset containing the embeddings $\mathbf{y}_i = E(t_i)$ of the textual labels $\mathbf{t}_i = C(\mathbf{t}_i, p)$. 

\subsection{Dimensionality Reduction and Visualization}
\label{sec:dr_vis}
Dimensionality reduction techniques are widely used to visualize high-dimensional data in a lower-dimensional space while preserving key data relationships. Classical methods such as Principal Component Analysis (PCA)\,\cite{peason1901lines,jolliffe1986principal} focus on linear transformations to retain variance. However, these do not scale well for datasets with a high intrinsic dimensionality. Subsequent techniques such as Multidimensional Scaling (MDS)\,\cite{torgerson58,kruskal1964multidimensional} and Isomap\,\cite{tenenbaum2000global} and its variations\,\cite{chen2006improved} aim to preserve distances between sample pairs in the data space $\mathbb{R}^n$ or on specific manifolds therein. However, such approaches do not work well when the data is not located on such manifolds. Neighborhood-preserving techniques like t-SNE\,\cite{maaten2008visualizing} and UMAP\,\cite{mcinnes2018umap} are particularly good when one aims to highlight clusters of similar samples present in the data. Several surveys cover the usage of DR methods for visual exploration. In particular, Nonato and Aupetit\,\cite{nonato18} discuss 28 such methods from the perspective of exploration tasks these support. Espadoto \textit{et al.}\,\cite{espadoto19} compare 44 the ability of 44 DR techniques to capture data structure, measured by seven quality metrics. Additional surveys cover DR methods based on the type of optimization technique they use when computing their 2D embedding\,\cite{maaten09_survey}, implementation techniques used\,\cite{sorzano14_survey,engel12}, or discuss specific classes of techniques such as linear\,\cite{cunningham15_survey} or non-linear ones\,\cite{yin07_survey}. We refer to these papers for a comprehensive overview of existing DR techniques.

More importantly for our goals, all DR techniques assume that the dimensions of their input datasets $D$ are able to capture the aspects they aim to depict, such as clusters of similar data samples. When this is not the case, additional methods known generally as \emph{supervised dimensionality reduction} can `guide' the projection output $P(D)$ to capture aspects not present directly in $D$ but which can be inferred from $D$\,\cite{espadoto2021SSNP,classimap,wang18}. 

\subsection{Semantic Alignment and Embedding Steering}
\label{sec:semantic}
Advances in vision and language models in the form of multimodal learning\,\cite{radford2021learning} have enabled text and image representations to be semantically aligned in a shared embedding space. Models such as CLIP\,\cite{radford2021learning} achieve this by learning a joint representation where textual descriptions act as class prototypes. This model also provides the ability of \emph{zero-shot classification}, \emph{i.e.,} classifying 
data whose types has not been seen during training.

Recent work has explored methods to \emph{steer} embeddings by leveraging the semantic alignment between image and text data; proposed prompt-driven knowledge distillation to improve model generalization\,\cite{li2024promptkd}. Separately, unsupervised prompt learning for vision-language models was explored\,\cite{huang2022unsupervised}. These works demonstrate that textual conditioning can significantly impact representation learning. Textual guidance was also used to improve classification quality by enhancing semantic consistency within embedding spaces through the fusion of textual and visual features\,\cite{visapp25}. By leveraging textual prompts related to the class labels, such as `This is digit 9' or `This is an animal,' embeddings can be structured to reflect higher-level semantic distinctions, leading to better classification performance. 

Beyond classification, embedding transformations and zero-shot classification have also been integrated into clustering workflows. This includes clustering multimodal data through learned cluster prototypes to improve separability in classification settings\,\cite{ma2024mode} and methods for incorporating zero-shot prompts into vision-language models to improve cluster coherence\,\cite{zeng2024modalprompt, chen2021multimodal}. These works emphasize the role of \emph{semantic alignment} in clustering.

While such works highlight the benefits of using aligned embeddings, there remains a gap in applying this type of semantic labeling to steer projections in an unsupervised manner. Existing DR techniques focus on exploring distances and neighborhoods. Supervised techniques expect class labels, which are relatively simple representations of semantic similarity. This raises the question of whether language-driven embeddings can offer a more interpretable and flexible complement to traditional projection techniques. We answer this question positively in our work as described next.

	\section{Method}
	\label{sec:method}
	We next introduce our approach for constructing steerable and improved 2D projections of high-dimensional data. Our method leverages zero-shot classification using Qwen\,\cite{bai2025qwen25vltechnicalreport} to generate semantic textual labels which are next fused with the original data dimensions before applying DR. Our method has the following steps (see also Fig.~\ref{fig:teaser}):

\smallskip
\noindent\textbf{Data Embedding:} For a given dataset $O$, generate embeddings $D$ using CLIP (Sec.~\ref{sec:embedding_extraction}); \\
\noindent\textbf{Labeling:} For each $\mathbf{x}_i \in D$, generate textual class labels $\mathbf{t}_i \in T$ by running zero-shot classification with Qwen based on a guiding textual prompt $p$ (Sec.~\ref{sec:zeroshot}); \\
\noindent\textbf{Label Embedding:} Generate embeddings $Y$ of the text labels $T$ using CLIP (Sec.~\ref{sec:embedding_labels}); \\
\noindent\textbf{Fusing Embeddings:} Fuse data and label embeddings $D$ and $Y$ to create a new dataset (Sec.~\ref{sec:fusion}); \\
\noindent\textbf{Projection:} Visualize the fused embeddings (Sec.~\ref{sec:dr}). \\

\subsection{Data Embedding}
\label{sec:embedding_extraction}
This step is quite straightforward: Given a text or image dataset $O$, we generate embeddings
\[
\mathbf{x}_i = E(\mathbf{o}_i)
\]
for all samples $\mathbf{o}_i \in O$ using the embedding model $E$ provided by CLIP. CLIP uses a dimensionality of $n=512$ for its embeddings, ensuring compatibility with its pretrained multimodal model. This step yields a dataset of embeddings $D = \{ \mathbf{x}_i \} \subset \mathbb{R}^{512}$.

\subsection{Labeling}
\label{sec:zeroshot}
To introduce semantic structure into the embedding space, we generate textual labels for each observation using zero-shot classification with Qwen. Labels are dynamically generated based on a so-called \emph{guiding prompt} which determines how the model describes the data. For image data, Qwen is prompted to classify the image based on structured queries. For text data, Qwen processes the document alongside a structured query to infer the category.

Given a classifier $C$ (Qwen in our case), an input observation $\mathbf{o}_i$, and the guiding prompt $p$, the textual label $\mathbf{t}_i$ is obtained as
\begin{equation}
	\mathbf{t}_i = C(\mathbf{o}_i, p)
\end{equation}
where $p$ is a guiding prompt, such as
\begin{itemize}
	\item \emph{"What is this? Answer: This is a <category>."}
	\item \emph{"What category does this belong to?"}
\end{itemize}
This results in a set of structured descriptions $T=\{\mathbf{t}_i\}$, such as
\begin{itemize}
	\item \emph{"This is digit 2 even"} (for an MNIST image);
	\item \emph{"This is a car, vehicle"} (for a CIFAR-10 image).
\end{itemize}

Since different guiding prompts influence how the data is described, they allow for alternative interpretations of the same dataset $O$ which, in turn, lead to different projections highlighting these interpretations.

\subsection{Label Embedding}
\label{sec:embedding_labels}
The textual labels $T$ are embedded using the same embedder $E$ as for the original data $O$ (Sec.~\ref{sec:embedding_extraction}), that is, by computing 
\begin{equation}
	\mathbf{y}_i = E(\mathbf{t}_i), \mathbf{t}_i \in T
\end{equation}
where $\mathbf{y}_i $ is the $n$-dimensional embedding of textual label $\mathbf{t}_i$ generated by $C$. The set $Y=\{\mathbf{y}_i\}$ of embedded textual labels will co-exist in the same embedding space as the set of data embeddings $D$. This shared embedding space is crucial for our method because it allows us to
\begin{itemize}
	\item Embed images and texts into a semantically aligned feature space, ensuring consistency across modalities;
	\item Ensure comparability between data points and their generated labels, allowing them to be fused for integrating meaningful descriptions into data;
	\item Enable flexible projection refinement by adjusting the guiding prompt, which allows for different perspectives on the dataset.
\end{itemize}

In other words, by leveraging this unified multimodal space, our method treats both images and text in a consistent manner, allowing projections to be structured based on the guiding prompt $p$ rather than being purely driven by the input data $O$.

\subsection{Fusing Embeddings}
\label{sec:fusion}
To incorporate semantic structure into the embedding space, we fuse the original data embeddings with the textual label embeddings. The fused representation is computed as
\begin{equation}
	\mathbf{x}' = \alpha \mathbf{x} + (1 - \alpha) \mathbf{y}, ~~\mathbf{x} \in D, ~~\mathbf{y} \in Y,
\label{eqn:alpha}
\end{equation}
where $\alpha$ is a fusion weight. Larger $\alpha$ values favor projections which highlight more the \emph{data} $D$; smaller values favor projections which emphasize more the \emph{textual labels} $Y$. We set $\alpha$ by default to 0.5 in our experiments. We denote the set of fused embeddings as $D' = \{\mathbf{x}'\}$. This fusion step ensures that projections (computed next) are structured not just by data similarity but by high-level semantic relationships, guided by the zero-shot classification process and guiding prompt.

\subsection{Projection}
\label{sec:dr}
Finally, we take the dataset $D'$ of fused embeddings, and project it with any suitable DR technique $P$, yielding the scatterplot $P(D')$.
Different projection techniques $P$ will leverage different aspects of the fused embedding (Eqn.~\ref{eqn:alpha}) as explained next in Sec.~\ref{sec:experiments}.

	\section{Experiments}
	\label{sec:experiments}
	  We conduct a series of experiments to evaluate the impact of guiding prompts (Sec.~\ref{sec:zeroshot}) on embedding visualization. Our analysis focuses on two main aspects: (1) assess whether the integration of semantic textual labels generated via zero-shot classification improves the structure and interpretability of projections; and (2) investigate how different types of guiding prompts influence data organization in the projected space. 

\begin{figure*}[htpb!]
\centering
\includegraphics[width=1.0\textwidth]{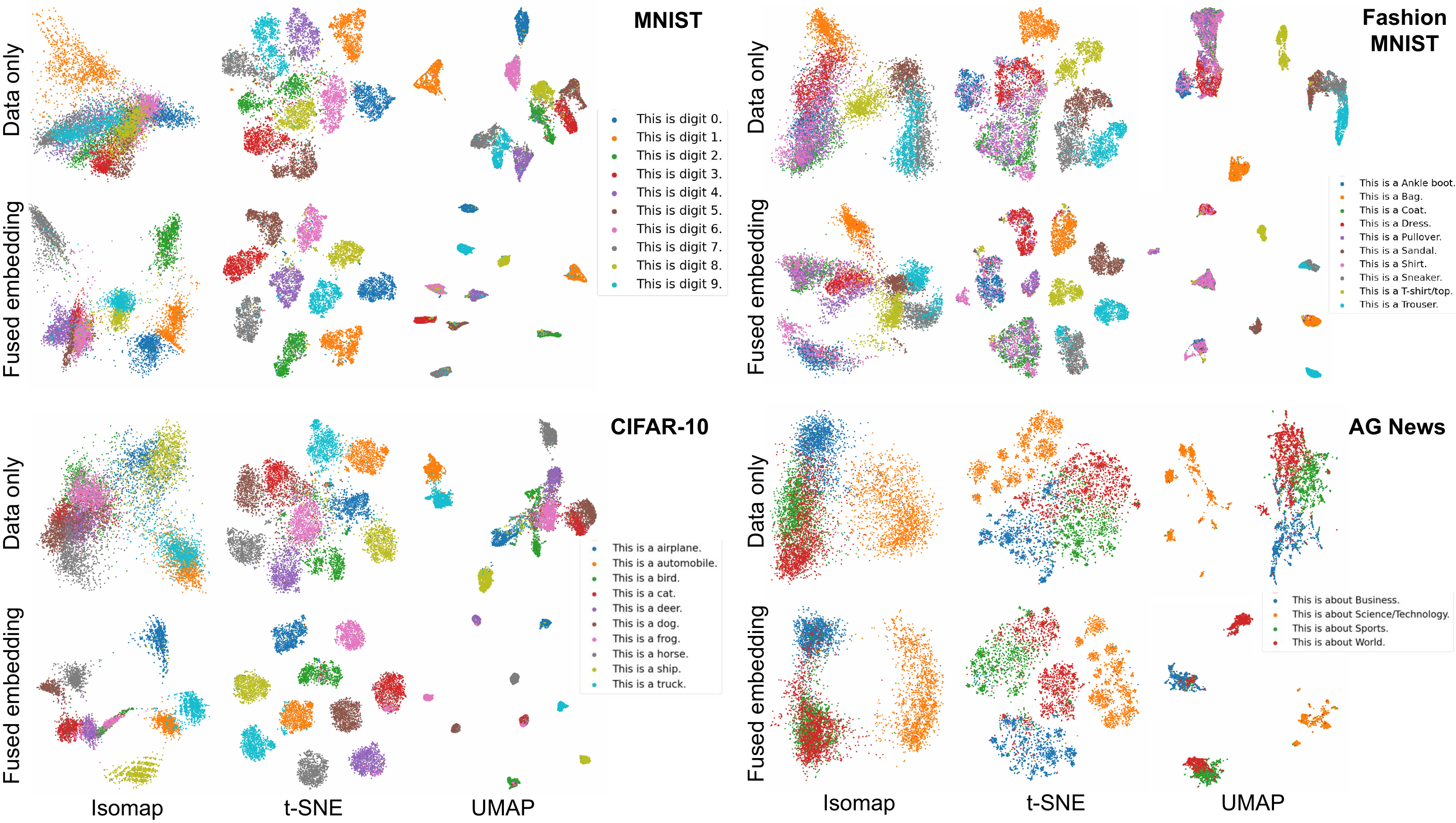} 
\caption{Experiment 1: Projections created for the four studied datasets using three different projection techniques. Top rows show projections $P(D)$ created using only the embedded data $D$. Bottom rows shows projections $P(D')$ created using fused embeddings based on zero-shot classification.}
\label{fig:experiment_closed}
\end{figure*}

Our approach is dataset-agnostic and can be applied to any text or image dataset. We selected four datasets with different characteristics that may influence aspects of our method, particularly in zero-shot classification prompts and embedding fusion strategies, as follows: \\

\smallskip
\noindent\textbf{AG News}\,\cite{zhang2015character}: A text dataset with four top categories (World, Sports, Business, Science/Technology), used to test our method on text data; \\
\noindent\textbf{MNIST}~\cite{lecun1998gradient}: Handwritten digits dataset with 10 classes, commonly used for benchmarking classification models; \\
\noindent\textbf{Fashion-MNIST}\,\cite{xiao2017fashion}: A dataset of clothing and footwear categories, designed as a more complex alternative to MNIST; \\
\noindent\textbf{CIFAR-10}\,\cite{krizhevsky2009learning}: An image dataset with 10 object categories, including animals and vehicles, introducing more visual complexity than Fashion-MNIST; \\


To visualize the high-dimensional fused embeddings, we use three well-established and commonly used DR techniques, namely, Isomap\,\cite{tenenbaum2000global}, t-SNE\,\cite{maaten2008visualizing}, and UMAP\,\cite{mcinnes2018umap}. Using each of these techniques, we compute two-dimensional scatterplots $P(D')$ for each of the four above-mentioned datasets, following the pipeline described in  Sec.~\ref{sec:method}. 

We describe next three experiments that evaluate our two earlier-listed aspects of interest (Secs.~\ref{sec:exp1}-\ref{sec:exp3}). Furthermore, we compute several quality metrics to gauge the projections computed by our technique (Sec.~\ref{sec:metrics}). All guiding prompts used for zero-shot classification are listed alongside the descrpition of the experiments.

\subsection{Experiment 1: Expected labels}
\label{sec:exp1}
In this experiment, we select guiding prompts based on the expected class labels that are known to exist in the dataset, \textit{e.g.,} digits from 0-9 in MNIST, pieces of clothing in Fashion-MNIST. To evaluate our method's steering ability, we generate two visualizations for each pair of projection technique $P$ and dataset $O$:

\smallskip
\noindent\textbf{Embedded data only}: Standard projection $P(D)$, used as baseline for comparison;\\
\noindent\textbf{Fused embeddings (data + label)}: Incorporate textual labels generated by Qwen using guiding prompts to introduce semantic structure into the projection.

For this experiment, we used the following guiding prompts:

\smallskip
\noindent \textbf{AG News}: \emph{"What is this news article about? Answer with the structure: This is about <World | Sports | Business | Science/Technology>."}; \\
\noindent \textbf{CIFAR-10}: \emph{"What is this? Answer with the structure: This is a <airplane | automobile | bird | cat | deer | dog | frog | horse | ship | truck>."}; \\
\noindent \textbf{Fashion-MNIST}: \emph{"What clothing item is this? Answer with the structure: This is a <T-shirt/top | Trouser | Pullover | Dress | Coat | Sandal | Shirt | Sneaker | Bag | Ankle boot>."}; \\
\noindent \textbf{MNIST}: \emph{"What digit is this? Answer with the structure: This is digit <$0 ~|~ 1 ~|~ \ldots 9$>."}; \\

Figure~\ref{fig:experiment_closed} shows the results of this experiment. In all cases (datasets and projection techniques), we see that the fused embeddings yield a sharper separation of clusters of same-label samples as compared to the baseline projections. In the same time, the overall `visual signature' of the underlying projection technique (Isomap, t-SNE, and UMAP) is kept -- Isomap shows clusters of various shapes and spreads; t-SNE exhibits its typical round, organic-shaped, clusters; and UMAP creates sharp, concentrated, clusters separated by large amounts of whitespace. 

A few things are important to mention here. First, since the prompts essentially ask for the same type of information as the ground-truth labels capture, \emph{e.g.}, whether a CIFAR-10 image is a dog, the fact that our fused projections align well with the ground-truth labels validates empirically a good execution of the zero-shot classification. Separately, one could argue that the same strong cluster separation we obtain by our method could be reached by adding ground-truth labels to the dimensions of the data $D$ to project -- essentially, constructing a supervised projection. However, this would only work for \emph{labeled} data; our method does not expect any such labels. Also, in this supervised projection scenario, ground-truth labels would bring in a single \emph{categorical} dimension to enrich the data. In contrast, our textual labels are mapped to $n=512$ \emph{continuous} dimensions (Sec.~\ref{sec:embedding_labels}). This extracts far more nuanced, and richer, information from the dataset that can next steer the projection. Finally, prompts can be easily \emph{changed} to reflect any user concern of interest that should steer the projection; ground-truth labels are, in contrast, typically fixed. Our next experiments further show the added-value of the flexible prompt engineering for steering the projection.

\subsection{Experiment 2: Hierarchical grouping}
\label{sec:exp2}
We extend the previous experiment by creating enriched projections that encode high-level semantic categories, to determine whether our method can steer projections to form hierarchies of clusters. For this, we used the following guiding prompts:

\smallskip
\noindent \textbf{CIFAR-10}: \textit{"What is this? Answer with the structure: This is a <airplane | automobile | bird | cat | deer | dog | frog | horse | ship | truck> <vehicle | animal>."}; \\
\noindent \textbf{Fashion-MNIST}: \textit{"What type of clothing item is this? Answer with the structure: This is a <T-shirt/top | Trouser | Pullover | Dress | Coat | Sandal | Shirt | Sneaker | Bag | Ankle boot> <tops | bottoms | outerwear | footwear | accessory>."}; \\
\noindent \textbf{MNIST}: \textit{"What digit is this? Answer with the structure: This is digit <digit from 0 to 9> <even | odd>."}; \\

We did not include the AG News dataset in the hierarchical grouping experiment because it has only four classes that do not group well semantically (Sports, World, Business, and Science). As an alternative, we could use generated labels to \emph{split} these four classes into subclasses. However, verifying that the resulting projections would bring added value is more complex since we do not have any ground-truth information on such subclasses. As such, we leave this type of experiment out of our current scope.

\begin{figure}[htpb!]
\centering
\includegraphics[width=0.5\textwidth]{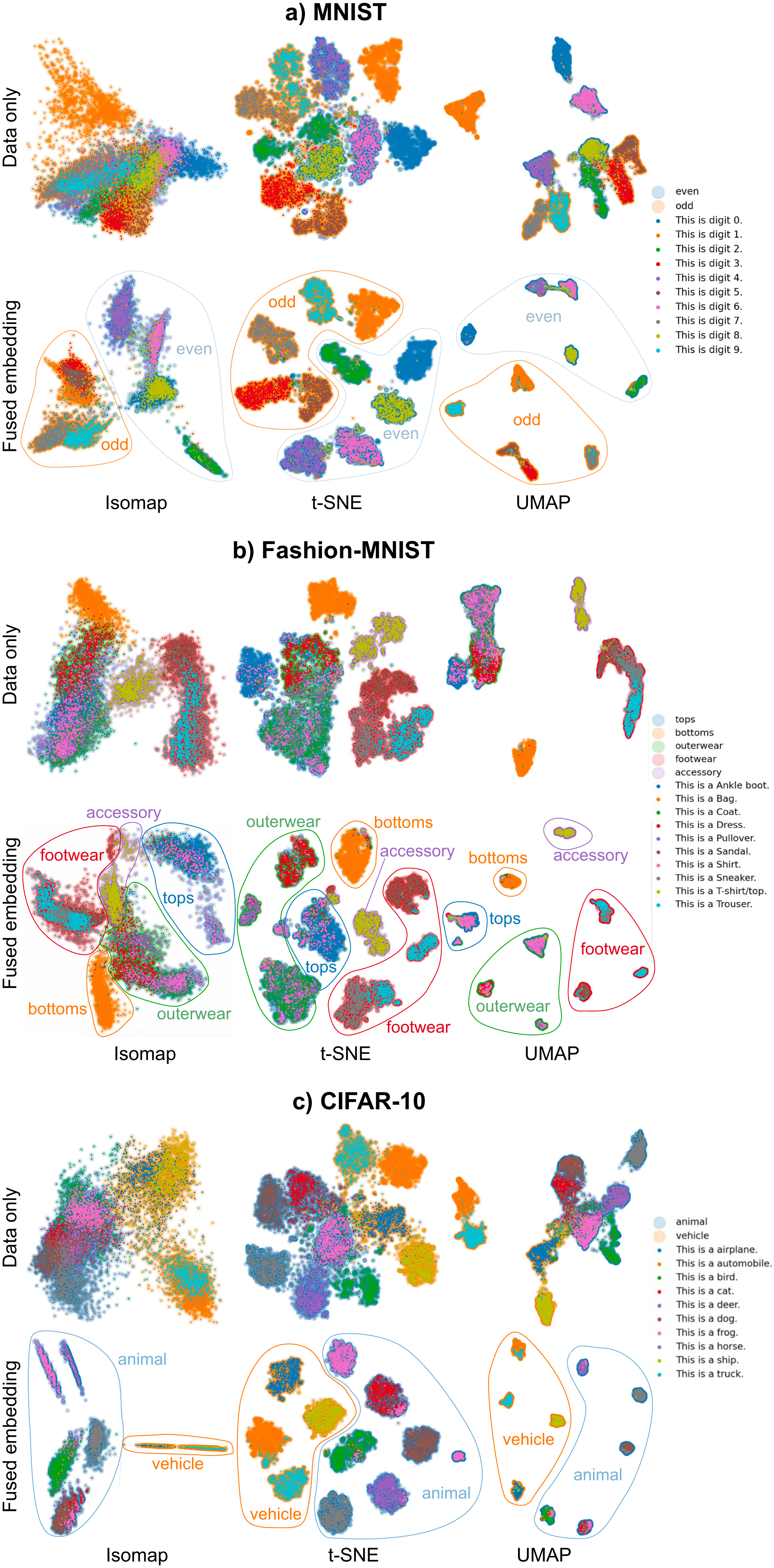}
\caption{Experiment 2: Projections created for MNIST (a), FashionMNIST (b), and CIFAR-10 (c), using Isomap, t-SNE, and UMAP, with data grouped by a higher-level class. Top row `data only' show projections created using only the embedded data. Bottom row shows projections created using the fused embeddings, where class labels contain a high-level grouping. The color of the larger dots represent the class of the grouping. The color of the smaller dots represent the class of the data point. Closed curves show approximate group boundaries.}
\vspace{-0.15cm}
\label{fig:exp2}
\end{figure}

Figure~\ref{fig:exp2} shows the results for this experiment. We encode the class label, \emph{e.g.}, \emph{airplane} for CIFAR, in the color of the scatterplot points. Separately, we encode the higher-level category extracted by our prompts, \emph{e.g.}, \emph{animal} for CIFAR, in the color of outlines drawn around each dot. When the underlying projection groups same-class samples, we see this in the usual way, as same-color clusters. Additionally, if samples in a cluster also have the same higher-level category values, we see this in the color of the band formed around the respective cluster. If we consider the baseline projections, we see that samples are grouped well with respect to their classes, as expected. However, we see that the resulting clusters are not arranged in any particular way with respect to each other. For example, in the MNIST projection, odd and even digit clusters are interspersed in the projection. In contrast, when we use our fused embeddings, points are still grouped to reflect label similarity (as expected) but, in addition, clusters having similar high-level category values are placed close to each other. For example, in the MNIST projection, the odd-digit clusters (having an orange outline) appear in the lower part of the projection); even-digit clusters (having a blue outline) are placed in the upper part of the projection.

\subsection{Experiment 3: Qualitative queries}
\label{sec:exp3}
In our first experiment we used prompts to basically \emph{replicate} the information reflected by the labels -- as the aim was to validate the generation of meaningful prompts, and since in this case the labels are known to us. The second experiment uses slightly more advanced prompts which essentially group the information reflected by the labels and showed that the resulting clusters get grouped accordingly in the projection. Still, one could argue that such prompts are relatively simple manipulations of the underlying data in ways that reflect some kind of knowledge related to the labels. If prompts were limited to such effects,  one could validly question their added value. 

We show the power of prompt-based projection steering by taking the idea of using a zero-shot classifier further. We now classify the data points based on attributes that are not explicitly encoded in the dataset in the form of class labels or any other variables. For this, we used the Fashion-MNIST and CIFAR-10 datasets and, for each one, we created new categorical groupings having two values each, as follows:

\smallskip
\noindent \textbf{Fashion-MNIST}: \textit{"Describe this clothing item with the structure: This is <Modern | Old-fashioned>."}; \\
\noindent \textbf{CIFAR-10}: \textit{"What time is it? Answer with the structure: It is <Day | Night>."}; \\

\begin{figure}[htpb!]
\centering
\includegraphics[width=0.47\textwidth]{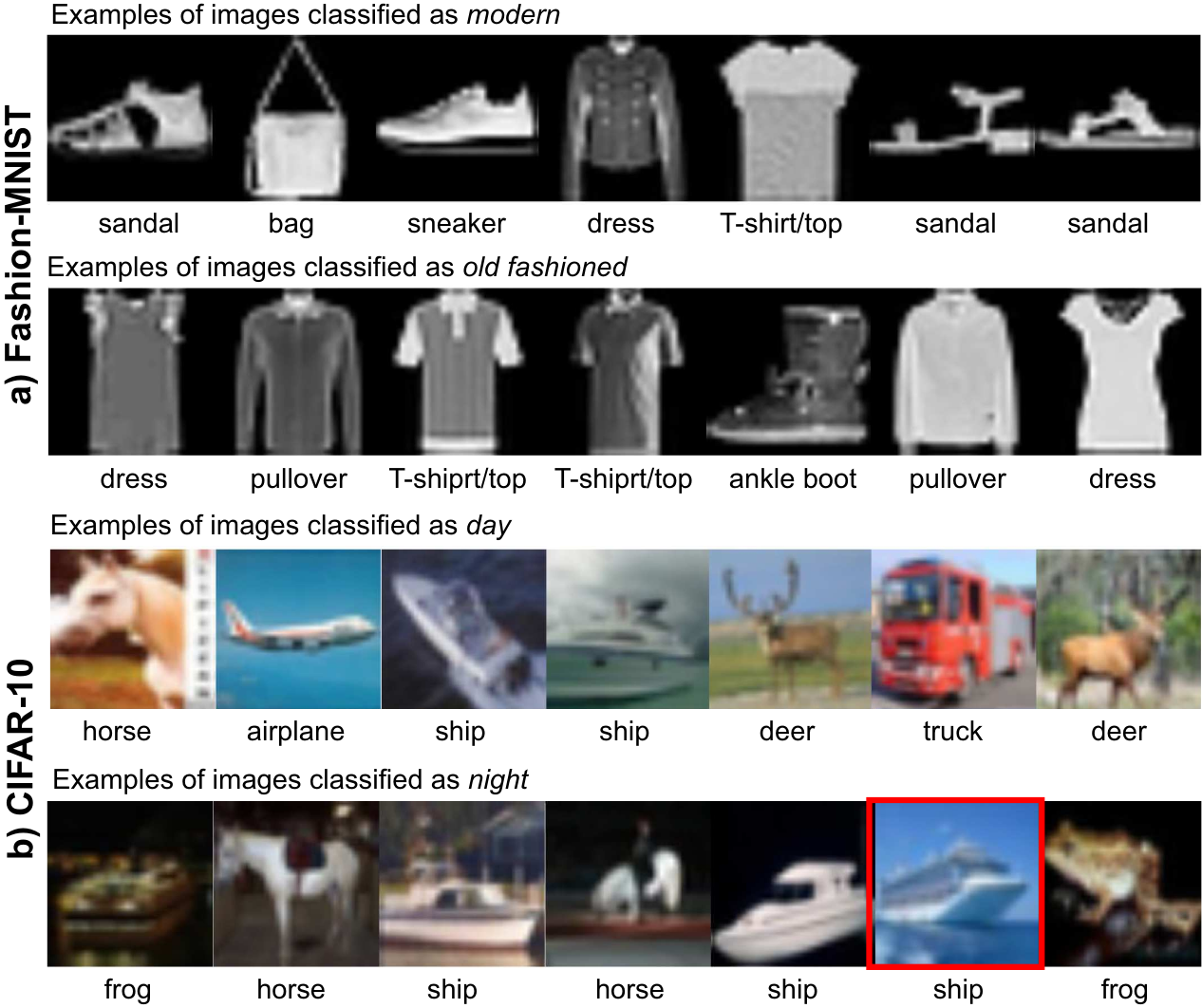}
\caption{Examples of input images from the Fashion-MNIST and CIFAR datasets with the inferred results by our zero-shot classification.}
\label{fig:zeroshot}
\end{figure}

\begin{figure*}[htpb!]
\centering
\includegraphics[width=0.9\textwidth]{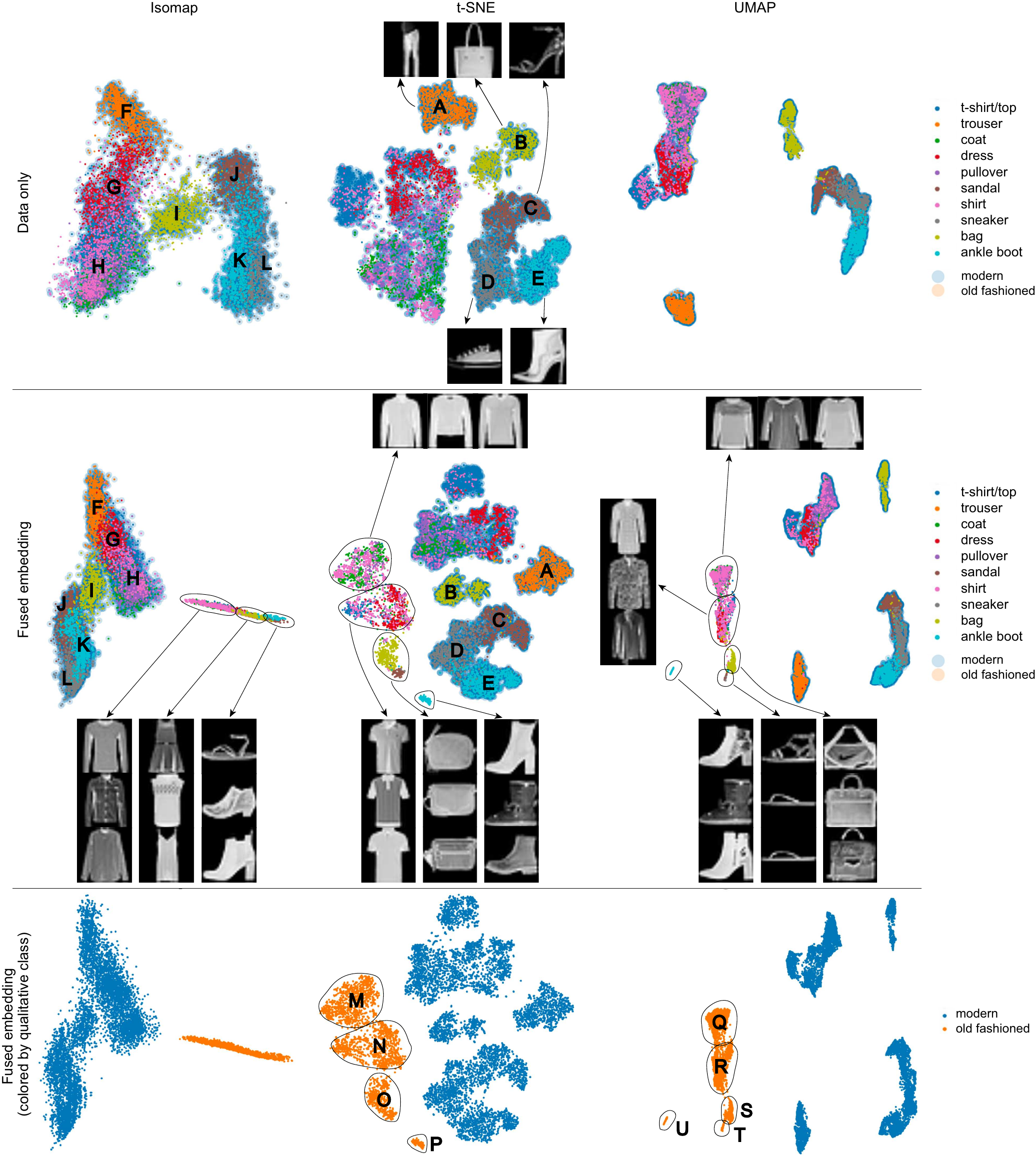}
\caption{Projections created for the FashionMNIST dataset using three techniques (Isomap, t-SNE, and UMAP), with the data grouped by a qualitative prompt. 
First row of projections (`data only'): Points projected without using the textual prompts, colored by class, with outline color indicating prompt-induced classes. Second row of projections (`fused embeddings'): Points projected using the data and textual prompts, colored as above. Third row of projections: Points projected as in the second row, colored only by prompt-induced classes. Thumbnails show selected images in various regions of the projections.}
\label{fig:exp3}
\end{figure*}

Figure~\ref{fig:zeroshot} shows examples of images with their original classes from Fashion-MNIST and CIFAR-10 as well as the qualitative classes inferred by the zero-shot classification using our prompts, namely (\emph{modern}, \emph{old fashioned}) for Fashion-MNIST, and (\emph{day}, \emph{night}) for CIFAR-10. While we have no ground-truth to formally verify this zero-shot classification -- and, recall, this cannot be done in general since the user is free to express \emph{any} concerns of interest in the textual prompts -- we see that the inferred qualitative labels correlate well with our human understanding of what the images show. However, this is not a watertight mechanism -- see, for example the image outlined in red for CIFAR-10. This shows a ship which, given the surrounding skylight, appears to be in an evening setting; as such, the \emph{night}  zero-shot classification for this image is likely incorrect.

\noindent\textbf{Fashion-MNIST results:} Figure~\ref{fig:exp3} compares the baseline and fused-embedding projections for the Fashion-MNIST dataset in this experiment. Similar to Fig.~\ref{fig:exp2}, we encode in the top two rows of Fig.~\ref{fig:exp3} the class of the data points in their colors and the class of the grouping induced by our prompts in the point outline colors, respectively. Figure~\ref{fig:exp3} (bottom row) shows points colored only by the prompt-induced classes, for additional clarity.

\begin{figure*}[htpb!]
\centering
\includegraphics[width=0.78\textwidth]{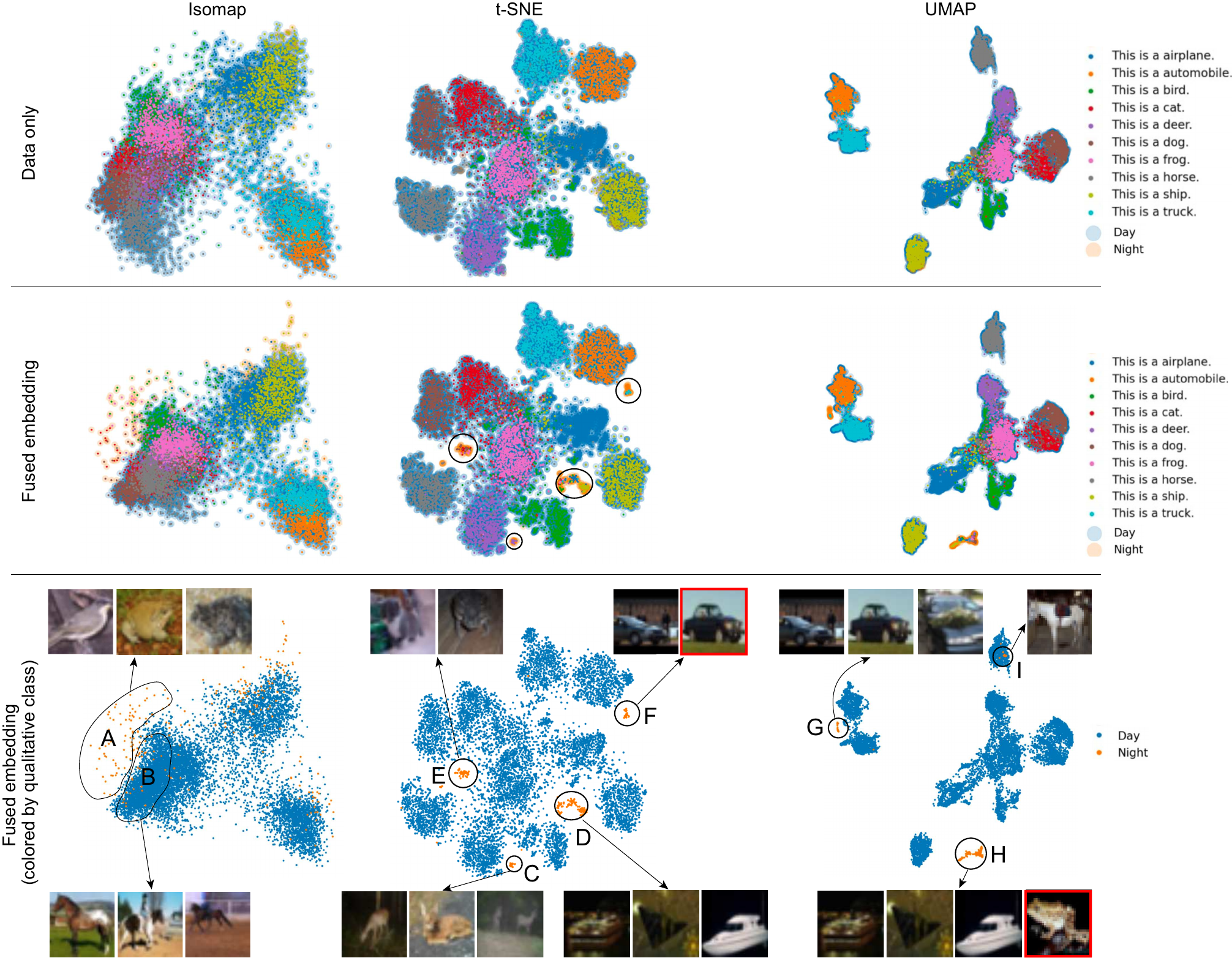}
\caption{Projections created for the CIFAR-10 dataset using Isomap, t-SNE, and UMAP, with data grouped by a qualitative prompt. 
First row (`data only'): Points projected without using the textual prompts, colored by class, with outline color indicating prompt-induced classes. Second row (`fused embeddings'): Points projected using the data and textual prompts, colored as above. Third row of projections: Points projected as in the second row, colored only by prompt-induced classes. Thumbnails show selected images in various regions of the projections.}
\label{fig:exp3_cifar}
\vspace{-5mm}
\end{figure*}

The differences between the fused-embedding projections and the baseline ones appear now more prominent than for the first two experiments: While prompts \emph{sharpened} the projection (experiment 1) and \emph{grouped} point clusters (experiment 2), they now \emph{reshape} the projection to reflect the aspects captured by the added attribute.
When using the fused embeddings, projections split into well-separated orange and blue set of clusters corresponding to the old-fashioned, respectively modern, clothing images -- see Fig.~\ref{fig:exp3} (bottom row). Also, we see that the fused projection does not unnecessarily alter areas in the baseline projection when not needed for its `extra task' of showing the information present in the textual labels. For example, the structure present in the baseline Isomap projection formed by the orange-red-purple clusters (Fig.~\ref{fig:exp3} top row F,G,H) is also visible in the steered projection (Fig.~\ref{fig:exp3} middle row F,G,H); the brown, cyan, and gray clusters (J,K,L) are close to each other in both the baseline and the steered projection; and cluster I (yellow) is roughly in the middle of the projection, between all other clusters. A similar situation appears when considering t-SNE: Both the baseline and the steered projection show a good separation of clusters A-E (with cluster A even keeping similar shape). The same happens for the baseline and steered UMAP projections. 

Let us now examine what is \emph{different} between the baseline and steered projections. When using Isomap as underlying technique, we see a salient elongated cluster split away from the bulk of the projection (Fig.~\ref{fig:exp3}, bottom row, orange points). Coloring points by class (Fig.~\ref{fig:exp3} middle row), we see that this cluster consists of three main groups of points, namely shirts (pink), dresses and bags (yellow and red), and ankle boots (cyan). The steered projection thus not only separated images classified as old fashioned, but kept their organization given by the other underlying attributes. 

When using t-SNE as underling technique, the separated orange points (Fig.~\ref{fig:exp3} bottom row) appear to form four sub-clusters (M-P in Fig.~\ref{fig:exp3}). We turn again to color coding to explain these sub-clusters -- see Fig.~\ref{fig:exp3} middle row. We find that these sub-clusters capture meaningful structure: pullovers and coats (pink and green, M); (t-)shirts, tops, and dresses (blue, pink, red, N); bags mostly (yellow, O); and a well-separated cluster of ankle boots (cyan, P). Note that the relative mix of certain item types, \emph{e.g.} (t)-shirts, tops, and dresses in cluster N, is \emph{not} due to the projection steering. These items were also hard to separate in the baseline projection (Fig.~\ref{fig:exp3} top row).

When using UMAP as underlying technique, the steering effects are quite similar: The baseline projection keeps roughly unchanged in areas where samples are not classified differently by the user prompt -- see the similar position and shape of the yellow, orange, brown, cyan, gray, and cyan clusters in Fig.~\ref{fig:exp3} top and middle rows. Note that these are precisely the clusters also kept similar between baseline and steered when using Isomap and t-SNE. For UMAP, the orange cluster created by the steered projection appears to consist of a larger structure having four sub-clusters (Q-T) and a separate small cluster (U) -- see Fig.~\ref{fig:exp3} bottom row. As earlier, we explain these structures by color coding: The clusters organize the images into shirts (Q), a mix of shirts and dresses (R), bags (S), sandals (T), and ankle boots (U).

\noindent\textbf{CIFAR-10 results:} Figure~\ref{fig:exp3_cifar} shows the results of experiment 3 for the CIFAR-10 dataset. As for the previous datasets, we start by examining what is \emph{different} between the baseline and steered projections. For this dataset, Fig.~\ref{fig:exp3_cifar} bottom row shows that the two classes inferred by our prompt are quite unbalanced -- there are far fewer night images (orange points) than daylight ones (blue points). This is in contrast with our prompting of Fashion-MNIST where the two qualitative classes were more comparable in size (see Fig.~\ref{fig:exp3} bottom row). As a consequence, the steered projection for CIFAR-10 does not need to change that much as compared to the baseline to accommodate the extra information given by the prompt. The structure of the baseline projection (without using qualitative prompts, top row in Fig.~\ref{fig:exp3_cifar}) is very much the same as the structure of the prompt-steered projection (middle row in Fig.~\ref{fig:exp3_cifar}). 

Yet, the prompt-steered projection succeeds in separating well the images having the two induced classes by the qualitative prompt, \emph{i.e.}, daytime ones (blue), respectively night ones (orange). The separation strength correlates with the clustering ability of the underlying projection (Fig.~\ref{fig:exp3}, CIFAR-10, bottom row): Isomap places the orange points on the periphery of the fused projection. The new features $Y$ created by the textual prompt extract differences between daylight and night images -- yet, these are not strong enough to make Isomap create separate clusters of orange points. This is in line with the fact that Isomap is a \emph{distance} preserving projection. Brushing the clusters A (orange points) and its neighbors B (daylight points) in the Isomap projection shows that the images A do not appear, indeed, to have been taken in daylight; in contrast, images B show all a clear daylight skyline. 

In contrast to Isomap, t-SNE separates the orange points in several well-delimited clusters which it then arranges within the already-existing structure of the baseline projection -- see C--F in Fig.~\ref{fig:exp3} CIFAR-10, t-SNE column. This is in line with the well-known ability of t-SNE to delineate \emph{neighborhoods} of similar samples. Brushing these clusters we find wild animals (C), boats (D), a mixed collection of cats, dogs, and frogs (E), and various automobiles (F).
All images show a night or twilight setting -- which the zero-shot classifier could interpret as night -- see for example the car outlined in red in inset F (Fig.~\ref{fig:exp3} bottom row). 
More interesting is the placement of these orange clusters within the overall projection -- see Fig.~\ref{fig:exp3_cifar}, t-SNE, middle row. We see that C is close to the purple (deer) cluster; D is between the green (bird and yellow (ship) clusters; E in the middle of the brown (deers), red (cats), cyan (frogs), and gray (horses) clusters; and F is close to the orange (automobiles) cluster. This placement is fully in line with our observation concerning the images gathered in these orange clusters by the steered projection.
  
Finally, UMAP shows an even stronger cluster-separation behavior -- in line with its well-known tendency to create compact point clusters separated by wide amounts of whitespace.
The two orange clusters it creates (G, H in Fig.~\ref{fig:exp3_cifar} bottom row) are, as for t-SNE, well separated from the blue points. Upon brushing these, we find that G contains basically the same automobile images as cluster F in the steered t-SNE projection; and H contains the boat images earlier seen in cluster D with t-SNE.
At this point, one wonder how this relates to the finer-grained separation of orange points -- the four different clusters we have seen with t-SNE. Upon closer inspection, we
find an orange cluster (marked I) embedded \emph{inside} the topmost cluster in Umap which groups horse images. Thes orange points are night images of horses -- see thumbnail in the figure. Separately, brushing the orange cluster H at the bottom of the projection, we discover that this merges night-like images of boats and of some animals, see \emph{e.g.} the frog thumbnail outlined in red. Simply put, the Umap projection merged the clusters D and E, which were depicted separated by t-SNE, into a single cluster H.

\begin{figure}[htpb!]
  \centering
  \includegraphics[width=0.48\textwidth]{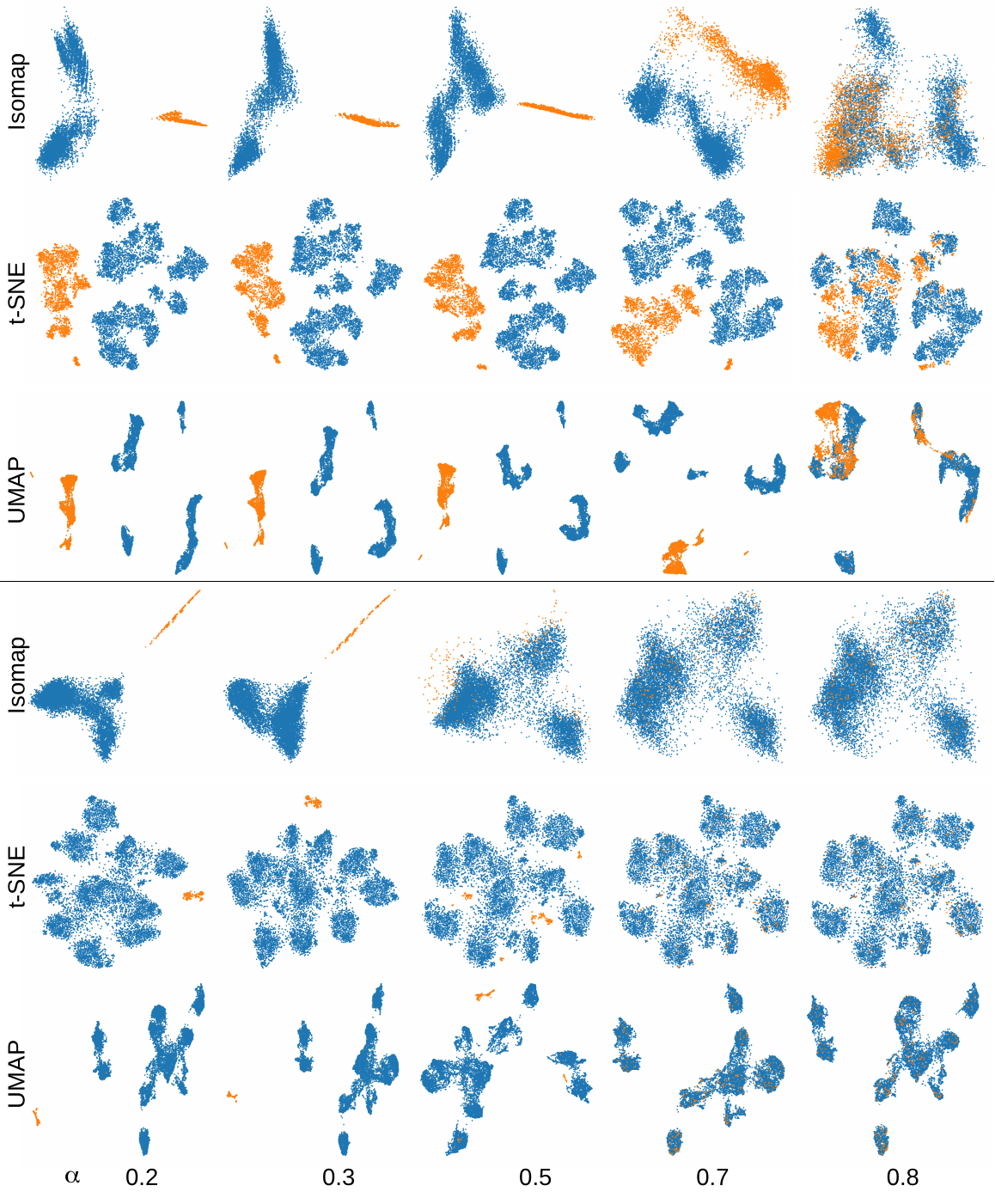}
  \caption{Projections created for FashionMNIST (top) and CIFAR-10 (bottom) using three projection techniques, varying the value of $\alpha$ from $0.2$ (emphasis on qualitative labels) to $0.8$ (emphasis on baseline data). Colors represent each qualitative class.}
  \label{fig:exp3_alpha}
  \vspace{-0.15cm}
\end{figure}

\noindent\textbf{Controlling the steering:} While Fig.~\ref{fig:exp3} shows how text prompts can be used to steer the projection, some users may find the changes between the baseline projection -- what they would be accustomed with -- and the steered one too abrupt. We can easily control this difference: By varying the weight $\alpha$ (Eqn.~\ref{eqn:alpha}), we smoothly morph between a purely data-driven projection ($\alpha=1$) and one which only emphasizes the textual labels ($\alpha=0$). Figure~\ref{fig:exp3_alpha} shows this effect. Note how, as $\alpha$ increases, the orange and blue points (encoding the values of textual labels) increasingly mix, since the projection uses increasingly less the textual label information. 

\subsection{Quality Metrics Assessment}
\label{sec:metrics}
We have showed that prompts can be used to steer the appearance of the projection based on user-defined concerns of interest. However, one important concern follows here: How faithful are such projections to reflecting the structure of the \emph{original} data, \emph{i.e.}, the dataset $D$?
If this structure is not reflected well, then fusing $D$ with the prompt embeddings $Y$ will not lead to a useful visualization.

To assess the effectiveness of embedding fusion for projection quality, we compute the following metrics (see also Tab.~\ref{tab:metrics}): 

\begin{table}[htbp!]
\centering
\caption{Projection quality metrics. Right column gives the metric ranges, with optimal values marked in bold.}
\label{tab:metrics}
\resizebox{\columnwidth}{!}{%
\begin{tabular}{ | l | c |  c |}
\hline
\textbf{Metric}         & \textbf{Definition}                                                                     & \textbf{Range} \\ \hline
Trustworthiness ($T$)   & $ 1 - \frac{2}{NK(2n-3K-1)}\sum_{i=1}^{N}{\sum_{j \in U^{(K)}_i}{(r(i,j) - K)}}$        & $[0, \mathbf{1}]$ \\
Continuity ($C$)        & $ 1 - \frac{2}{NK(2n-3K-1)}\sum_{i=1}^{N}{\sum_{j \in V^{(K)}_i}{(\hat{r}(i,j) - K)}}$  & $[0, \mathbf{1}]$ \\
Shepard diagram corr. ($R$) & $\rho$ of $(\|\mathbf{x}_i-\mathbf{x}_j\|, \|P(\mathbf{x}_i)-P(\mathbf{x}_j)\|), 1 \leq i \leq N, i \neq j$ &  $[0,\mathbf{1}]$ \\
Silhouette score ($S$)  & $ \frac{b - a}{max(a, b)}                                                               $ & $[-1, \mathbf{1}]$ \\
\hline
\end{tabular}}
\end{table}

\noindent\textbf{Trustworthiness ($T$)\,\cite{venna_visualizing_2006}:} Measures the fraction of close points in $D$ that are also close in $P(D)$. $T$ tells how much one can trust that local patterns in a projection $P(D)$ represent actual patterns in the data $D$. In the definition (Tab.~\ref{tab:metrics}), $U^{(K)}_i$ are the $K$ nearest neighbors of point $i$ in the 2D space which are not among the $K$ nearest neighbors of point $i$ in $\mathbb{R}^n$; and $r(i, j)$ is the rank of the 2D point $j$ in the ordered-set of nearest neighbors of $i$ in 2D. We choose $K=7$, in line with\,\cite{maaten_dimensionality_2009,martins_explaining_2015,espadoto19}.

\noindent\textbf{Continuity ($C$)\,\cite{venna_visualizing_2006}:} Measures the fraction of close points in $P(D)$ that are also close in $D$. In the definition (Tab.~\ref{tab:metrics}), $V^{(K)}_i$ are the $K$ nearest neighbors of point $i$ in $\mathbb{R}^n$ which are not among the $K$ nearest neighbors in 2D; and $\hat{r}(i, j)$ is the rank of the $\mathbb{R}^n$ point $j$ in the ordered set of nearest neighbors of $i$ in $\mathbb{R}^n$. As with $T$, we choose $K=7$.

\noindent\textbf{Shepard diagram correlation ($R$)\,\cite{joia_local_2011}:} The Shepard diagram is a scatter plot of the pairwise distances between all points in $P(D)$ \textit{vs} the corresponding distances in $D$. The closer the plot is to the main diagonal, the better overall distance preservation is. Plot areas below, respectively above, the diagonal show distance \emph{ranges} for which false neighbors, respectively missing neighbors, occur. We measure how close a Shepard diagram is to the ideal main diagonal line by computing its Spearman rank correlation $\rho$. A value of $R=1$ indicates a perfect (positive) correlation of distances.

\noindent\textbf{Silhouette score ($S$)\,\cite{ROUSSEEUW198753}:} Quantifies cluster cohesion and separation by comparing each cluster's tightness and separation, respectively defined as the mean distance between a sample ($a$) and all other points in the same class, and the mean distance between a sample and all other points in the next nearest cluster ($b$). A value of $S=1$ indicates a very well-separated clustering, whereas $S=0$ indicates overlapping clusters. A value of $S=-1$ indicates an incorrect clustering. 

Table~\ref{tab:exp1} shows the values of the above four metrics for the experiments 1 to 3 described earlier in this section. In these tables, the labels `Data' indicate the values for the baseline projections $P(D)$; labels `Combined' indicate the values for our fused projections $P(D')$. The first observation is that the values for our fused projections are very close to the baseline values. This indicates that our projections reflect the actual data structure in $D$ at basically the same level as the baseline projections. More interestingly, in most cases the metric values for our fused projections are actually \emph{larger} than the baseline ones, so our fused projections are actually \emph{better} at preserving data structure. Let us analyze what this means: Our fused dataset $D'$ augments the baseline data $D$ with extra attributes $Y$ which come, via the prompts, also from an analysis of $D$. Projecting this entire space $D'$ is found easier by the considered techniques (Isomap, t-SNE, UMAP) than projecting the original space $D$. The attributes $Y$ appear to perform a `disentanglement' of $D$, \emph{i.e.}, their addition makes $D$ easier to project.

\begin{table}[htpb]
\centering
\caption{Metrics computed for experiments 1 to 3 (Secs.~\ref{sec:exp1}-\ref{sec:exp3}). Numbers in bold indicate best measurement in the group.}
\label{tab:exp1}
\resizebox{0.9\columnwidth}{!}{%

  \begin{tabular}{|l|l|l|l|l|l|l|l|}
  \hline
  \multirow{25}{*}{\rotatebox{90}{Experiment 1}} & Dataset                        & Projection              & Type     & T               & C               & R               & S               \\ \cline{2-8} 
                                 & \multirow{6}{*}{AG News}       & \multirow{2}{*}{Isomap} & Data     & 0.7789          & \textbf{0.9550} & 0.4780          & 0.2861          \\ \cline{4-8} 
                                 &                                &                         & Combined & \textbf{0.8355} & \textbf{0.9550} & \textbf{0.7147} & \textbf{0.2879} \\ \cline{3-8} 
                                 &                                & \multirow{2}{*}{T-SNE}  & Data     & 0.9810          & 0.9691          & 0.3820          & 0.2497          \\ \cline{4-8} 
                                 &                                &                         & Combined & \textbf{0.9897} & \textbf{0.9820} & \textbf{0.5112} & \textbf{0.2868} \\ \cline{3-8} 
                                 &                                & \multirow{2}{*}{UMAP}   & Data     & 0.9558          & 0.9726          & 0.4283          & 0.3106          \\ \cline{4-8} 
                                 &                                &                         & Combined & \textbf{0.9749} & \textbf{0.9851} & \textbf{0.6234} & \textbf{0.4310} \\ \cline{2-8} 
                                 & \multirow{6}{*}{CIFAR-10}      & \multirow{2}{*}{Isomap} & Data     & 0.8363          & 0.9530          & \textbf{0.6492} & 0.0456          \\ \cline{4-8} 
                                 &                                &                         & Combined & \textbf{0.9505} & \textbf{0.9717} & 0.6485          & \textbf{0.4555} \\ \cline{3-8} 
                                 &                                & \multirow{2}{*}{T-SNE}  & Data     & 0.9873          & 0.9744          & 0.5460          & 0.3687          \\ \cline{4-8} 
                                 &                                &                         & Combined & \textbf{0.9937} & \textbf{0.9905} & \textbf{0.5579} & \textbf{0.5498} \\ \cline{3-8} 
                                 &                                & \multirow{2}{*}{UMAP}   & Data     & 0.9639          & 0.9762          & \textbf{0.6010} & 0.3870          \\ \cline{4-8} 
                                 &                                &                         & Combined & \textbf{0.9820} & \textbf{0.9916} & 0.4755          & \textbf{0.7265} \\ \cline{2-8} 
                                 & \multirow{6}{*}{Fashion-MNIST} & \multirow{2}{*}{Isomap} & Data     & 0.8963          & 0.9719          & 0.6809          & 0.1489          \\ \cline{4-8} 
                                 &                                &                         & Combined & \textbf{0.9495} & \textbf{0.9828} & \textbf{0.7085} & \textbf{0.2136} \\ \cline{3-8} 
                                 &                                & \multirow{2}{*}{T-SNE}  & Data     & 0.9925          & 0.9840          & 0.5254          & 0.2342          \\ \cline{4-8} 
                                 &                                &                         & Combined & \textbf{0.9955} & \textbf{0.9925} & \textbf{0.5551} & \textbf{0.2398} \\ \cline{3-8} 
                                 &                                & \multirow{2}{*}{UMAP}   & Data     & 0.9715          & 0.9863          & \textbf{0.5779} & 0.2901          \\ \cline{4-8} 
                                 &                                &                         & Combined & \textbf{0.9854} & \textbf{0.9942} & 0.4175          & \textbf{0.3371} \\ \cline{2-8} 
                                 & \multirow{6}{*}{MNIST}         & \multirow{2}{*}{Isomap} & Data     & 0.8299          & 0.9661          & \textbf{0.6243} & 0.0406          \\ \cline{4-8} 
                                 &                                &                         & Combined & \textbf{0.9073} & \textbf{0.9749} & 0.6232          & \textbf{0.3011} \\ \cline{3-8} 
                                 &                                & \multirow{2}{*}{T-SNE}  & Data     & 0.9889          & 0.9786          & 0.4507          & 0.3533          \\ \cline{4-8} 
                                 &                                &                         & Combined & \textbf{0.9948} & \textbf{0.9924} & \textbf{0.5328} & \textbf{0.5124} \\ \cline{3-8} 
                                 &                                & \multirow{2}{*}{UMAP}   & Data     & 0.9657          & 0.9798          & 0.4310          & 0.4811          \\ \cline{4-8} 
                                 &                                &                         & Combined & \textbf{0.9828} & \textbf{0.9930} & \textbf{0.5824} & \textbf{0.6951} \\ \hline
  \end{tabular}}
\resizebox{0.9\columnwidth}{!}{%
\begin{tabular}{|l|l|l|l|l|l|l|l|}
\hline
\multirow{19}{*}{\rotatebox{90}{Experiment 2}} & Dataset                        & Projection              & Type     & T               & C               & R               & S               \\ \cline{2-8} 
                                & \multirow{6}{*}{CIFAR-10}      & \multirow{2}{*}{Isomap} & Data     & 0.8363          & \textbf{0.9530} & 0.6492          & 0.0456          \\ \cline{4-8} 
                                &                                &                         & Combined & \textbf{0.9484} & 0.9520          & \textbf{0.8010} & \textbf{0.2799} \\ \cline{3-8} 
                                &                                & \multirow{2}{*}{T-SNE}  & Data     & 0.9873          & 0.9744          & \textbf{0.5464} & 0.3684          \\ \cline{4-8} 
                                &                                &                         & Combined & \textbf{0.9937} & \textbf{0.9905} & 0.5331          & \textbf{0.5200} \\ \cline{3-8} 
                                &                                & \multirow{2}{*}{UMAP}   & Data     & 0.9639          & 0.9778          & \textbf{0.5750} & 0.4065          \\ \cline{4-8} 
                                &                                &                         & Combined & \textbf{0.9821} & \textbf{0.9916} & 0.3776          & \textbf{0.6747} \\ \cline{2-8} 
                                & \multirow{6}{*}{Fashion-MNIST} & \multirow{2}{*}{Isomap} & Data     & 0.8963          & 0.9719          & \textbf{0.6809} & \textbf{0.1489} \\ \cline{4-8} 
                                &                                &                         & Combined & \textbf{0.9236} & \textbf{0.9768} & 0.6552          & 0.1114          \\ \cline{3-8} 
                                &                                & \multirow{2}{*}{T-SNE}  & Data     & 0.9925          & 0.9840          & 0.5268          & 0.2346          \\ \cline{4-8} 
                                &                                &                         & Combined & \textbf{0.9955} & \textbf{0.9925} & \textbf{0.5464} & \textbf{0.2617} \\ \cline{3-8} 
                                &                                & \multirow{2}{*}{UMAP}   & Data     & 0.9718          & 0.9862          & \textbf{0.5806} & 0.2858          \\ \cline{4-8} 
                                &                                &                         & Combined & \textbf{0.9855} & \textbf{0.9942} & 0.5141          & \textbf{0.3063} \\ \cline{2-8} 
                                & \multirow{6}{*}{MNIST}         & \multirow{2}{*}{Isomap} & Data     & 0.8300          & 0.9661          & \textbf{0.6243} & 0.0406          \\ \cline{4-8} 
                                &                                &                         & Combined & \textbf{0.8976} & \textbf{0.9762} & 0.4696          & \textbf{0.2616} \\ \cline{3-8} 
                                &                                & \multirow{2}{*}{T-SNE}  & Data     & 0.9889          & 0.9786          & \textbf{0.4508} & 0.3525          \\ \cline{4-8} 
                                &                                &                         & Combined & \textbf{0.9950} & \textbf{0.9926} & 0.4397          & \textbf{0.5026} \\ \cline{3-8} 
                                &                                & \multirow{2}{*}{UMAP}   & Data     & 0.9660          & 0.9797          & \textbf{0.4386} & 0.4990          \\ \cline{4-8} 
                                &                                &                         & Combined & \textbf{0.9845} & \textbf{0.9936} & 0.3442          & \textbf{0.6735} \\ \hline
\end{tabular}}

\resizebox{0.9\columnwidth}{!}{%

  \begin{tabular}{|l|l|l|l|l|l|l|l|}
  \hline
  \multirow{13}{*}{\rotatebox{90}{Experiment 3}} & Dataset                        & Projection              & Type     & T               & C               & R               & S                \\ \cline{2-8} 
                                 & \multirow{6}{*}{CIFAR-10}      & \multirow{2}{*}{Isomap} & Data     & 0.7780          & \textbf{0.9360} & 0.6620          & \textbf{-0.1286} \\ \cline{4-8} 
                                 &                                &                         & Combined & \textbf{0.7823} & 0.9300          & \textbf{0.6938} & -0.1443          \\ \cline{3-8} 
                                 &                                & \multirow{2}{*}{T-SNE}  & Data     & 0.9537          & \textbf{0.9273} & 0.6200          & -0.1267          \\ \cline{4-8} 
                                 &                                &                         & Combined & \textbf{0.9570} & 0.9257          & \textbf{0.6273} & \textbf{-0.1176} \\ \cline{3-8} 
                                 &                                & \multirow{2}{*}{UMAP}   & Data     & 0.8528          & 0.9431          & 0.6251          & \textbf{-0.1180} \\ \cline{4-8} 
                                 &                                &                         & Combined & \textbf{0.8585} & \textbf{0.9458} & \textbf{0.6745} & -0.1247          \\ \cline{2-8} 
                                 & \multirow{6}{*}{Fashion-MNIST} & \multirow{2}{*}{Isomap} & Data     & 0.8963          & 0.9719          & 0.6809          & \textbf{0.1489}  \\ \cline{4-8} 
                                 &                                &                         & Combined & \textbf{0.9217} & \textbf{0.9753} & \textbf{0.8670} & -0.0651          \\ \cline{3-8} 
                                 &                                & \multirow{2}{*}{T-SNE}  & Data     & 0.9925          & 0.9840          & 0.5254          & \textbf{0.2348}  \\ \cline{4-8} 
                                 &                                &                         & Combined & \textbf{0.9945} & \textbf{0.9902} & \textbf{0.5856} & 0.1281           \\ \cline{3-8} 
                                 &                                & \multirow{2}{*}{UMAP}   & Data     & 0.9720          & 0.9861          & 0.5729          & \textbf{0.2822}  \\ \cline{4-8} 
                                 &                                &                         & Combined & \textbf{0.9817} & \textbf{0.9917} & \textbf{0.7152} & 0.0506           \\ \hline
  \end{tabular}}
\end{table}
\subsection{Implementation}
\label{sec:time}
All experiments discussed above were run on a dual-socket Intel Xeon Silver 4310 (48 threads) and 512 GB of RAM, and an NVidia RTX A6000 with 48 GB VRAM. Our implementation and datsasets will be made publicly available after peer review.

\noindent\textbf{Scalability:} Since zero-shot classification is a key part of our method, we measured the time needed to classify different numbers of samples (500, 1000, 2000, 3000, 5000) from the CIFAR-10 dataset using Qwen for the three types of strategies used in our experiments (Secs~\ref{sec:exp1}-\ref{sec:exp3}). We found that all scale \emph{linearly} in the sample count: 2.4 images/second (hierarchical grouping, experiment 2); 2.9 images/second (expected labels, experiment 1); and 3.6 images/second (qualitative labels, experiment 3). While not real-time, such values demonstrate the applicability of our approach in explorative data visualization.

	\section{Discussion}
	\label{sec:discussion}
	We discuss next several key aspects of our method.

\noindent\textbf{Added value:} Our three experiments show different use-cases where our steered projections yield added value. In the simplest case, we provide a \emph{better separation} of clusters formed by data samples (Fig.~\ref{fig:experiment_closed}), also reflected in its slightly higher quality metrics as compared to baseline projections (Tab.~\ref{tab:exp1}). Such better separation is, arguably, one of the fundamental goals of many projection algorithms, and serves a wide range of applications. 
Importantly, we achieve this better separation \emph{without} supervision and/or the need of additional ground-truth labels -- the CLIP and Qwen engines we use are already \emph{pre-trained} for their respective tasks. 

Secondly, we provide a way to \emph{structure} a projection by positioning clusters of related samples close to each other (Fig.~\ref{fig:exp2}). This goes beyond the abilities of existing neighborhood-based projection techniques such as t-SNE or UMAP which only optimize for intra-cluster similarity. We offer a simple to use mechanism that enables users to create hierarchy-like structures of point clusters in a projection based on aspects they deem of interest (supplied via prompts). 

Finally, we take further this grouping idea by proposing guiding prompts with qualitative attributes -- that is, attributes which are in no way directly inferable from the data or label dimensions present in a dataset (Figs.~\ref{fig:exp3},~\ref{fig:exp3_cifar}). This allows users to freely steer a projection to reflect any question they have about the data, as well as smoothly control how much the projection should reflect their question \emph{vs} reflect the actual attributes present in the input dataset. Simply put, a projection is no longer a \emph{fixed} view on high-dimensional data -- but rather a view that adapts to both the data and the user's questions about the data.

\noindent\textbf{Limitations:} Our approach relies on the quality of zero-shot classification which, as shown, is not perfect (Fig.~\ref{fig:zeroshot}). Yet, as such classifiers improve, we can use them directly in our projection steering approach at no extra cost or technical difficulty. Moreover, the key idea behind our proposal is that users will pose \emph{multiple} questions to examine a given dataset. As such, the obtained overall insights should not be massively affected by the presence of a small number of wrong zero-shot labels. 
Separately, while our approach has linear time cost in the sample count, our current overall throughput is relatively slow -- tens of minutes are needed to generate the text embeddings $Y$ for a dataset of several thousand samples. However, once this process is done, navigating between the baseline and steered projection (Eqn.~\ref{eqn:alpha}) e done in real time. Moreover, pre-computing answers to a rich set of domain-specific questions for a given dataset collection can be easily done to increase speed.
Finally, our method currently only handles text and image data. Extending it to handle any other type of data is an important direction to consider next.

	\section{Conclusion}
	\label{sec:conclusion}
	We have presented a new method for creating user-steered projections of high-dimensional text and image data based on interactive semantic mapping. In contrast to existing projection techniques, which provide a \emph{single} view on the data (which typically fails to account for all patterns in the data), we allow users to create an open-ended set of  views by \emph{steering} the projection to emphasize aspects of interest provided via textual prompts. Users can next smoothly navigate between a traditional, purely data-based, projection, and one which emphasizes the aspects of interest captured in the prompts. To achieve this, we leverage CLIP's capability of semantic alignment for fusing embeddings of data and text labels, and also Qwen's capability of zero-shot classification for image data. Our method is simple to implement, generically handles any text and image datasets, and is simple to use -- one has to only specify a set of questions to be answered on the data (via textual prompts).

Many directions of future work exist. Our current textual prompts could be refined towards more natural-language-like text so users truly `get into a conversation' with the projection system, asking it to show the data from this or that perspective, and receiving the corresponding views. The same mechanism could be used to allow users to select a group of points in the projection and freely ask the system to explain, in natural language, what those samples have in common (`tell me what is this data'); or what makes several sample groups different. On the technical side, improved zero-shot classifiers and optimized GPU implementations would increase the accuracy of our proposal and also lower its execution time towards near-real-time performance. Last but most challenging, extending our proposal of creating user-steered projection for \emph{any} high-dimensional dataset (beyond text and images) is an objective we aim to further explore.

	\section*{Acknowledgments}
	This work was funded partially by FAPESP project 2022/15304-4 and MCTI (law 8.248, PPI-Softex - TIC 13 - 01245.010222/2022-44).
	
	\bibliographystyle{plain}
	\bibliography{refs}
	
\end{document}